# Anwendung von Causal-Discovery-Algorithmen zur Root-Cause-Analyse in der Fahrzeugmontage

Lucas Poßner[1], Lukas Bahr[2], Leonard Röhl[2], Christoph Wehner[3], Sophie Gröger[4]

[1] Montage im Werk Leipzig, BMW Group
[2] Digitalisierungsabteilung für die Batterieproduktion, BMW Group
[3] Lehrstuhl für kognitive Systeme, Universität Bamberg
[4] Professur für Fertigungsmesstechnik, TU Chemnitz
`Lucas.Possner@bmw.de`

**Abstract.** Root-Cause-Analyse (RCA) ist eine Methode des Qualitätsmanagements, die darauf abzielt, die Ursache-Wirkungs-Zusammenhänge von Problemen und deren zugrundeliegenden Ursachen systematisch zu untersuchen und zu identifizieren. Traditionelle Methoden basieren auf der Analyse von Problemen durch Fachexperten. In modernen Produktionsprozessen werden große Mengen an Daten erfasst. Aus diesem Grund werden zunehmend rechnergestützte und datengetriebene Methoden für die RCA verwendet. Eine dieser Methoden sind Causal-Discovery-Algorithmen (CDA). In dieser Veröffentlichung wird die Anwendung von CDA auf Daten aus der Montage eines führenden Automobilherstellers demonstriert. Die verwendeten Algorithmen lernen die kausale Struktur zwischen den Eigenschaften der hergestellten Fahrzeuge, der Ergonomie und des zeitlichen Umfangs der beteiligten Montageprozesse und qualitätsrelevanten Produktmerkmalen anhand repräsentativer Daten. Diese Publikation vergleicht verschiedene CDA hinsichtlich ihrer Eignung im Kontext des Qualitätsmanagements. Dafür werden die von den Algorithmen gelernten kausalen Strukturen sowie deren Laufzeit verglichen. Die vorliegende Veröffentlichung liefert einen Beitrag zum Qualitätsmanagement und demonstriert, wie CDA zur RCA in Montageprozessen eingesetzt werden können.

## 1  Einleitung

Trotz des technologischen Fortschritts und der zunehmenden Automatisierung in der Produktion bleibt die Rolle der Mitarbeiter in der Montage entscheidend [1]. In einer Fahrzeugendmontage werden etwa 95 % der Montageprozesse manuell von Mitarbeitern ausgeführt [2]. Dabei sind die Mitarbeiter kognitiven und physischen Belastungen ausgesetzt, welche ihre Leistungsfähigkeit beeinträchtigen können [3]. Hierbei besteht ein signifikanter Zusammenhang zwischen Ergonomie am Arbeitsplatz und der Produktqualität [4]–[6].

Eine hohe Produktqualität ist nicht nur ein wesentlicher Wettbewerbsvorteil, sondern auch ein entscheidender Faktor für den langfristigen Erfolg eines Unternehmens [7]–[9]. Zur Sicherstellung einer hohen Produktqualität ist ein effektives Qualitätsmanagement wichtig. Ein wichtiges Instrument im Qualitätsmanagement ist die Root-Cause-Analyse (RCA) [10]. RCA ist eine Methode, die darauf abzielt, die Ursache-Wirkungs-Zusammenhänge von Problemen und deren zugrundeliegenden Ursachen systematisch zu untersuchen und zu identifizieren. Traditionelle Methoden der RCA basieren auf der manuellen Untersuchung der Produktionsprozesse durch Prozessexperten [11]–[14]. Diese Ansätze stoßen jedoch angesichts der wachsenden Menge an in modernen Produktionsprozessen erhobenen Daten an ihre Grenzen [15]–[18]. Aus diesem Grund gewinnen datengetriebene Methoden zur RCA zunehmend an Bedeutung [15], [19]. Causal-Discovery-Algorithmen (CDA) sind ein Ansatz zur Durchführung von RCA. CDA lernen kausale Strukturen aus Daten und ermöglichen die Entdeckung der Ursachen von Problemen [20], [21].

In dieser Publikation soll untersucht werden, wie RCA in der Montage von Fahrzeugen mittels CDA durchgeführt werden kann. Zunächst wird in Kapitel 2 ein Überblick über den Stand der Technik gegeben. Danach werden in Kapitel 3 die in der Veröffentlichung genutzten CDA vorgestellt, sowie die Verknüpfung und Vorverarbeitung vorliegender Daten beschrieben. Danach werden die CDA für einen Benchmark auf simulierte Daten angewendet. Anschließend werden in Kapitel 4 die CDA mittels verschiedener Metriken sowie Laufzeit miteinander verglichen. Abschließend werden in Kapitel 5 die Ergebnisse diskutiert.

## 2 Stand der Technik

Folgendes Kapitel beschreibt den Stand der Technik für Causal-Discovery in Montageprozessen. Kapitel 2.1 beschreibt den Prozess der Montage von Fahrzeugen sowie den Einfluss von Ergonomie auf die Montageprozesse und erläutert den Zusammenhang mit der Produktqualität. Kapitel 2.2 führt die RCA ein und motiviert die Verwendung von rechnergestützten und datengetriebenen Methoden. Abschließend stellt Kapitel 2.3 die in der Veröffentlichung genutzten CDA vor.

### 2.1 Montage von Fahrzeugen

Als Montage bezeichnet man die Gesamtheit aller Vorgänge für den Zusammenbau von Körpern mit geometrisch bestimmter Form. Dazu gehören: Fügen, Handhaben, Prüfen, Justieren sowie Hilfsoperationen [22]. Die Komplexität der Montage im produzierenden Gewerbe wächst stetig. Gründe hierfür sind die steigende funktionale Komplexität der Produkte, eine erhöhte Anzahl miniaturisierter Fügeteile sowie eine zunehmende Variation der Produkte [23]. In dieser Veröffentlichung wird die manuelle Montage von Fahrzeugen betrachtet. Die manuelle Montage von Fahrzeugen umfasst den Prozess von der Einsteuerung der lackierten Karosse bis zur Übergabe des montierten, geprüften und in Betrieb genommenen Fahrzeugs an die Distributionslogistik. Sie unterscheidet sich gegenüber anderen Montageprozessen in Bezug auf die Anzahl der produzierten Einheiten, den Arbeitsumfang, die Größe des Liefernetzwerks, der Größe des Werkes und die Variabilität der Produkte sowie die Anzahl der benötigten Mitarbeiter [24]. Die Grundstruktur eines Montageprozesses von Fahrzeugen ist in Abb.1. dargestellt.

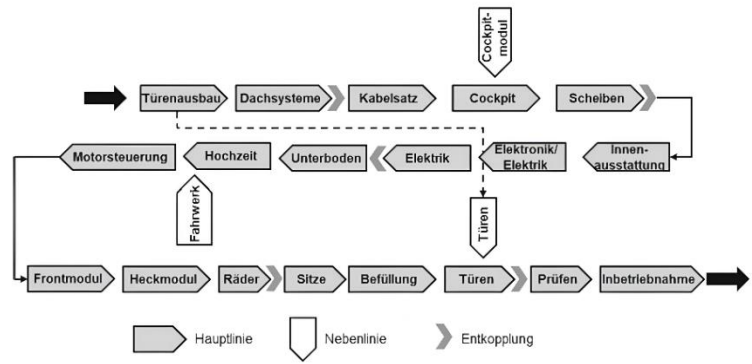

**Abb. 1.** Grundstruktur eines Montageprozesses von Fahrzeugen [24]. Bis zur Inbetriebnahme durchläuft das Fahrzeug eine komplexe Kette manueller Montageschritte.

Trotz des technologischen Fortschritts und der zunehmenden Automatisierung von Produktionsprozessen wird die Rolle der Mitarbeiter in der Montage entscheidend bleiben und sich kontinuierlich weiterentwickeln [1]. Dies liegt nicht zuletzt an dem teilweise geringem Automatisierungsgrad. In einer Fahrzeugendmontage liegt dieser beispielsweise bei rund 5 % [2]. Damit werden etwa 95 % der Montageprozesse durch Mitarbeiter ausgeführt. Dabei sind die Mitarbeiter nicht nur kognitiven, sondern auch physischen Belastungen ausgesetzt. Diese können wiederum die kognitive Leistungsfähigkeit erheblich beeinträchtigen [3]. Die Folgen von physischer und kognitiver Überlastung sind weitreichend und können zu einer verminderten Leistung der Mitarbeiter führen. Es ist zudem erwiesen, dass ein signifikanter Zusammenhang zwischen von Mitarbeitern verursachten Fehlern und der Ergonomie von deren Arbeitsplätzen besteht [4]. Die Bedeutung von Ergonomie in Montageprozessen umfasst daher nicht nur das Wohlbefinden und die Gesundheit der Mitarbeiter, sondern beeinflusst auch direkt die Produktqualität sowie die Produktivität [5]. Die Ergonomie von Arbeitsplätzen und Produktionsprozessen ist daher entscheidend für die Realisierung einer nachhaltigen Produktion [6].

### 2.2 Root-Cause-Analyse

Produktqualität ist nicht nur ein wesentlicher Wettbewerbsvorteil, sondern auch ein entscheidender Faktor für den langfristigen Erfolg eines Unternehmens [7]–[9]. Sie wird durch den Grad definiert, in dem ein Satz inhärenter Merkmale eines Produktes die gestellten Anforderungen erfüllt [25]. Um eine hohe Produktqualität sicherzustellen und fortlaufend zu optimieren, ist die RCA eine wichtige Methode im

Qualitätsmanagement [10]. Die Methode zielt darauf ab, Probleme und ihre zugrunde liegenden Ursachen systematisch zu untersuchen und zu identifizieren [26]. Effektive RCA erfordert eine tiefgehende Untersuchung der Produktionsprozesse [27]. Hierbei versteht man unter Produktionsprozessen einen Satz von in Wechselbeziehung oder Wechselwirkung stehenden Tätigkeiten, welche Eingaben in Ergebnisse umwandeln [25].

Traditionelle RCA Methoden basieren auf der manuellen Untersuchung der Produktionsprozesse durch Prozessexperten [11]–[14]. Die Qualität und Aussagekraft der Ergebnisse hängen dabei von den Fähigkeiten und dem Urteilsvermögen der analysierenden Personen ab [15]. Darüber hinaus können diese Methoden mit einem erheblichen zeitlichen Aufwand verbunden sein [11], [15]. Neuartige Methoden nutzen zudem Sprachmodelle, um auf Basis von Qualitätsdaten Schlüsse über Problemen zugrunde liegenden Ursachen zu ziehen [28]. In Produktionsprozessen wächst die Menge der erhobenen Daten, was die manuelle Untersuchung durch Prozessexperten an ihre Grenzen bringt [15]–[18]. Rechnergestützte und datengetriebene Methoden nutzen Produktionsdaten und haben das Potenzial, die Effizienz und die Genauigkeit der RCA zu verbessern [15], [19].

Eine dieser Methoden sind CDA. CDA lernen die kausale Struktur zwischen Zufallsvariablen als gerichteten azyklischen Graph (engl. Directed Acyclic Graph, Abk. DAG). Ein DAG besteht aus Knoten, die durch gerichtete Kanten verbunden sind und bei deren Durchlauf kein Knoten mehr als einmal durchlaufen wird. Dabei modellieren die Knoten die Zufallsvariablen und die Kanten die kausale Struktur [20], [21]. Dieser Ansatz wird in existierenden Veröffentlichungen bereits in verschiedenen Bereichen angewendet. So werden CDA beispielsweise in den Neurowissenschaften eingesetzt [29], in der Instandhaltung [11], [30], sowie bei der Untersuchung von Ausfällen in Cloud Services [31]. Auch in der Produktion von Elektrofahrzeugen gibt es erste Versuche, CDA anzuwenden [32].

## 2.3 Causal-Discovery-Algorithmen

CDA lernen Relationen zwischen Zufallsvariablen $X = (X_1, X_2, \dots X_N)$ anhand einer Menge von unabhängigen und identisch verteilten Stichproben $S = \{(x_{11}, x_{12} \dots x_{1N}), (x_{21}, x_{22} \dots x_{2N}) \dots, (x_{M1}, x_{M2} \dots x_{MN})\}$. Die Relationen bilden die kausale Struktur zwischen den Zufallsvariablen ab. Die Zufallsvariablen und Relationen können als DAG modelliert werden. Ein DAG $G$ ist ein Tupel $G = (V, E)$ von Knoten (Engl. Vertices) $V$ und Kanten (Engl. Edges) $E$ bei denen jede Kante $e = v_i \rightarrow v_j$ gerichtet ist und es keinen Pfad $v_1 \rightarrow v_2 \rightarrow \cdots \rightarrow v_L$ gibt, bei dem ein Knoten $v_i$ mehr als einmal durchlaufen wird. Dabei werden die Zufallsvariablen $X = (X_1, X_2, \dots X_N)$ als Knoten und die Relationen als gerichtete Kanten $e = v_i \rightarrow v_j$ modelliert [20], [21]. Die Anzahl möglicher gerichteter azyklischer Graphen (DAG) wächst exponentiell mit der Anzahl der Knoten [21]. Ein DAG ist in Abb. 2 dargestellt.

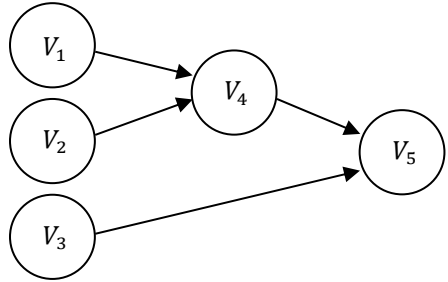

**Abb. 2.** Darstellung eines DAG. Dieser kann zur Modellierung der kausalen Struktur zwischen Zufallsvariablen verwendet werden.

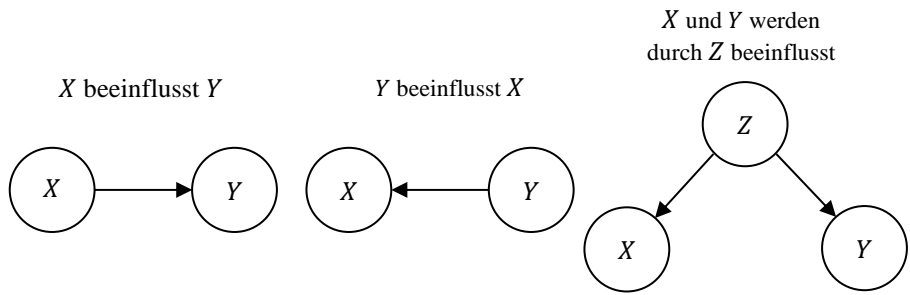

**Abb. 3.** Darstellung von kausalen Zusammenhängen, die eine Korrelation der Variablen $X$ und $Y$ implizieren. Links: $X$ beeinflusst $Y$. Mitte: $Y$ beeinflusst $X$. Rechts: $X$ und $Y$ werden durch $Z$ beeinflusst

Ein kausaler Zusammenhang zweier Zufallsvariablen $X$ und $Y$ impliziert eine statistische Korrelation. Diese lässt sich beispielsweise mittels des Pearson-Korrelationskoeffizienten von $X$ und $Y$ bestimmen [21]. Im Gegensatz dazu, ist die kausale Struktur nicht allein mittels $X$ und $Y$ bestimmbar, da verschiedene kausale Zusammenhänge zwischen $X$ und $Y$ die gleiche statistische Korrelation implizieren [20], [21]. Besteht zwischen zwei Zufallsvariablen $X$ und $Y$ eine statistische Korrelation, dann trifft eine der in Abb. 3 dargestellten Fälle zu [33]. Die kausale Struktur kann durch Hinzunahme von weiteren Variablen oder durch die Einbeziehung von domänenspezifischem Fachwissen bestimmt werden [26].

In der Literatur werden zwei Arten von CDA beschrieben. Die einen basieren auf diskreter Optimierung, die anderen auf kontinuierlicher Optimierung [34]. CDA lernen die Adjazenzmatrix $A \in \mathbb{R}^{d \times d}$. Dies ist eine $d \times d$ Matrix, welche einen Graphen mit $d$ Knoten repräsentiert. Ihre Zeilen und Spalten entsprechen den Knoten im Graphen und der Wert an der Position $(i,j)$ ist ein Maß für die Existenz sowie die Gewichtung der Kante von Knoten $i$ zu Knoten $j$. Die auf diskreter Optimierung basierenden CDA lösen folgendes Optimierungsproblem [35]:

$$\min_{A \in \mathbb{R}^{d \times d}} S(A) \\ \mathcal{G}(A) \in \text{DAG} \tag{1}$$

Die Verlustfunktion $S(A)$ wird unter der Nebenbedingung minimiert, dass die Adjazenzmatrix $A$ einen DAG repräsentiert. In [35] wird gezeigt, dass dies auch als ein kontinuierliches Optimierungsproblem ausgedrückt werden kann:

$$\min_{A \in \mathbb{R}^{d \times d}} S(A)$$
$$h(A) = 0 \qquad (2)$$

Dabei wird die Verlustfunktion $S(A)$ unter der Nebenbedingung $h(A) = 0$ minimiert. Die Funktion $h(A)$ erfüllt genau dann die Bedingung $h(A) = 0$, wenn die Adjazenzmatrix $A$ einen DAG repräsentiert. Die Funktionen $S(A)$ und $h(A)$ sind differenzierbar können mittels kontinuierlichen Optimierungsalgorithmen gelöst werden [35].

## 3 Methoden

Folgendes Kapitel stellt die Methoden für die Beantwortung der Forschungsfrage vor. Kapitel 3.1 gibt Überblick über die in der Veröffentlichung genutzten CDA. In Kapitel 3.2 wird die Struktur der vorliegenden Daten beschrieben. Danach wird in Kapitel 3.3 die Simulation der Daten für einen Benchmark der CDA beschrieben. Abschließend werden in Kapitel 3.4 die Metriken beschrieben, mittels derer die CDA verglichen werden sollen.

### 3.1 Causal-Discovery-Algorithmen

Der PC-Algorithmus [36] ist ein kombinatorischer, einschränkungsbasierter CDA, der diskreter Optimierung zum Lernen von DAGs nutzt. Dazu werden in einem vollständig ungerichteten Graphen paarweise Unabhängigkeitstests durchgeführt, um zu überprüfen, ob eine kausale Beziehung vorliegt. Kanten, die dies nicht erfüllen, werden entfernt. Anschließend wird die Orientierung der Kanten festgestellt.

Der NO TEARS-Algorithmus [35] ist der erste Algorithmus, der DAG mithilfe kontinuierlicher Optimierung lernt. Die Azyklizitätsbedingung ist genau dann erfüllt, wenn die Adjazenzmatrix $A$ den Eigenwert Null aufweist. Anstatt die Eigenwerte während der Optimierung neu zu berechnen, schlagen die Autoren vor, die Spur der Exponentialfunktion des Hadamard-Produkts der Adjazenzmatrix $A$ zu optimieren. Es ergibt sich für die Azyklizitätsbedingung

$$h(A) = \operatorname{tr}(\exp(A \circ A)) - d \qquad (3)$$

Ist die Azyklizitätsbedingung erfüllt, ist $h(A) = 0$; andernfalls ist $h(A) > 0$. Die Funktion ist differenzierbar und ermöglicht die Anwendung von kontinuierlicher Optimierung.

Mit DAGMA [37], führen die Autoren eine neue Azyklizitätsbedingung ein. Dafür wird die Adjazenzmatrix $A$ als M-Matrix repräsentiert und die Azyklizitätsbedingung wird durch die Maximierung der negativen Log Determinante berechnet. Die Autoren zeigen, dass insbesondere bei Graphen mit vielen Knoten, DAMGA bessere numerische Eigenschaften aufweist und schneller berechenbar ist.

## 3.2 Vorliegende Daten

**Struktur der Daten**. In der vorliegenden Veröffentlichung werden Daten aus der manuellen Montage von Fahrzeugen eines führenden Automobilherstellers betrachtet. Die Fahrzeuge werden in Taktfertigung montiert. Unter Taktfertigung versteht man eine Art der Fertigung, bei der die Arbeitsschritte in zeitlich gleich ausgelegte Takte aufgeteilt sind [38]. Die Arbeitsschritte werden im Folgenden als Teilvorgänge bezeichnet.

Aufgrund der unterschiedlichen Fahrzeugeigenschaften, wie Modell, Motor, Ausfuhrland sowie Sonderausstattungen, ergibt sich eine Variabilität der Teilvorgänge für jedes Fahrzeug und jeden Takt. Jedem Teilvorgang ist eine Bewertung bezüglich der Ergonomie sowie der veranschlagte Planzeit zugeordnet. Aus der Variabilität der Teilvorgänge ergeben sich für jedes Fahrzeug und für jeden Takt unterschiedliche Bewertungen bezüglich der Ergonomie sowie veranschlagte Planzeiten. Dies ist in Abb. 4 dargestellt.

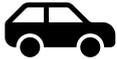

Fahrzeug 1
- Fahrzeugeigenschaft: 1; 2; 3
- Teilvorgang: 1; 2; 3; 4; 5; 6
- Ergonomie: 18; 32; 28; 40; 45; 31
- Planzeit: 1,3; 2,0; 1,4; 2,2; 1,4; 1,6

Fahrzeug 2
- Fahrzeugeigenschaft: 2; 3; 4
- Teilvorgang: 3; 4; 5; 6; 7; 8
- Ergonomie: 28; 40; 45; 31; 22; 31
- Planzeit: 1,4; 2,2; 1,4; 1,6; 2,4; 1,2

**Abb. 4.** Fahrzeugeigenschaften und Teilvorgänge sowie Bewertungen bezüglich deren Ergonomie und veranschlagten Planzeiten. Zusammengehörige Werte haben die gleiche Farbe.

Alle während der Montage verursachten Fehler werden in einem Qualitätsmanagementsystem dokumentiert und im Laufe der Montage nachgearbeitet. Dabei werden, unter anderem, die Art des Fehlers sowie der Takt dokumentiert.

**Verknüpfung der Daten.** In dieser Publikation werden durch Mitarbeiter verursachte Verschraubungsfehler beim Verbau einer Komponente betrachtet. Es werden relationale Daten aus zwei Quellen verknüpft. Die Spalten dieser Daten sowie die Art ihrer Attribute sind in Tabelle 1 und Tabelle 2 beschrieben.

**Tabelle 1.** Montierte Fahrzeuge und deren Eigenschaften.

| Spalte | Beschreibung | Variable |
|---|---|---|
| **Fahrzeugeigenschaften** (FaE) | Eigenschaften des Fahrzeugs | Liste, Nominal |
| **Fehler** (Fe) | Auftreten des betrachteten Fehlers | Binär |

**Tabelle 2.** Teilvorgänge und korrespondierende Fahrzeugeigenschaften.

| Spalte | Beschreibung | Variable |
|---|---|---|
| Fahrzeugeigenschaften | Korrespondierende Eigenschaften des Fahrzeugs | Liste, Nominal |
| **Ergonomie** (er) | Bewertung bezüglich der Ergonomie des Teilvorgangs | Positiv, Metrisch |
| **Planzeit** (pz) | Veranschlagte Planzeit des Teilvorgangs | Positiv, Metrisch |

Jedes Tupel in Tabelle 1 (Fahrzeugeigenschaften, Fehler) entspricht einem Fahrzeug und jedes Tupel in Tabelle 2 (Fahrzeugeigenschaften, Ergonomie, Planzeit) einem Teilvorgang. Die Eigenschaften der Fahrzeuge aus Tabelle 1 werden mit den zu dem Teilvorgang korrespondierenden Eigenschaften der Fahrzeuge aus Tabelle 2 verglichen. Wenn diese gleich sind, dann wird der entsprechende Teilvorgang bei der Montage von diesem Fahrzeug an dem betrachteten Takt durchgeführt. Jedem Fahrzeug $i$ aus Tabelle 1 wird eine Menge mit $L_i$ Elementen der für die Montage von diesem an dem betrachteten Takt durchgeführten Teilvorgänge sowie Bewertungen bezüglich deren Ergonomie und für diese veranschlagte Planzeiten zugeordnet Dies ist in Abb. 5 dargestellt.

Montiertes Fahrzeug und dessen Eigenschaften

Teilvorgänge und korrespondierende Fahrzeugeigenschaften

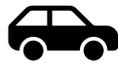
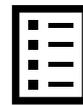

- Fahrzeugeigenschaft: 1; 2
- Teilvorgang: 1; 2; 3; 4
- Ergonomie: 18; 32; 28; 40
- Planzeit: 1,3; 2,0; 1,4; 2,2
- Fehler: Ja

- Fahrzeugeigenschaft: 1; 2
- Teilvorgang: 1; 2; 3; 4
- Ergonomie: 18; 32; 28; 40
- Planzeit: 1,3; 2,0; 1,4; 2,2

**Abb. 5.** Zuordnung der Fahrzeugeigenschaften zu Teilvorgängen sowie Bewertungen bezüglich deren Ergonomie und veranschlagten Planzeiten. Zusammengehörige Werte haben die gleiche Farbe.

Für jedes Fahrzeug wird der arithmetische Mittelwert der Bewertungen bezüglich der Ergonomie berechnet:

$$\text{Er}_i = \frac{1}{L_i} \sum_{j=1}^{L_i} \text{er}_{ij} \qquad (4)$$

Außerdem wird für jedes Fahrzeug die Summe der veranschlagten Planzeiten berechnet:

$$\text{Pz}_i = \sum_{j=1}^{L_i} \text{pz}_{ij} \tag{5}$$

Die nominalen Eigenschaften des Fahrzeugs werden jeweils in einer Spalte erfasst. Das Vorhandensein der Eigenschaft wird binär kodiert. Der arithmetische Mittelwert der Bewertung bezüglich der Ergonomie sowie die Summe der Planzeiten werden in jeweils 4 gleich große Intervalle unterteilt. Diese Intervalle werden jeweils in einer Spalte erfasst. Die Lage der Werte in dem entsprechenden Intervall wird binär kodiert. Die verarbeiteten Spalten sind in Tabelle 1 und Tabelle 2 markiert. Damit ergeben sich relationale Daten mit ausschließlich binären Attributen. Diese sind in Tabelle 3 dargestellt.

**Tabelle 3.** Relationale Daten mit binären Attributen, FaE: Fahrzeugeigenschaften, Er: Ergonomie, Pz: Planzeit, Fe: Fehler.

| FaE 1 | ... | FaE $K$ | Er Intervall 1 | ... | Er Intervall 4 | Pz Intervall 1 | ... | Pz Intervall 4 | Fe |
|---|---|---|---|---|---|---|---|---|---|
| Wahr | ... | Falsch | Wahr | ... | Falsch | Falsch | ... | Wahr | Wahr |
| Falsch | ... | Wahr | Falsch | ... | Wahr | Falsch | ... | Wahr | Falsch |
| Wahr | ... | Wahr | Wahr | ... | Falsch | Wahr | ... | Falsch | Wahr |
| Wahr | ... | Falsch | Wahr | ... | Falsch | Wahr | ... | Falsch | Falsch |
| Wahr | ... | Wahr | Falsch | ... | Wahr | Wahr | ... | Falsch | Wahr |

**Vorverarbeitung der Daten.** Die resultierenden Daten umfassen $N = 679$ Attribute und $M = 96.555$ Tupel. Die Laufzeit der verwendeten CDA ist von der Anzahl der Attribute sowie der Anzahl der Tupel der verwendeten Datensätze abhängig. Um die Laufzeit zu reduzieren, wurden die Daten vorverarbeitet. Dazu wurde die Annahme getroffen, dass die FaE 1 bis $K$ einen kausalen Zusammenhang mit der Er Intervall 1 bis 4 sowie der Pz Intervall 1 bis 4 und diese einen kausalen Zusammenhang mit dem Fehler aufweisen. Damit besteht die kausale Struktur aus drei aufeinanderfolgende Subgraphen $c_1, c_3, c_3$. Diese haben jeweils nur vorwärts gerichteten Kanten. Die kausale Struktur ist in Abb. 6 dargestellt.

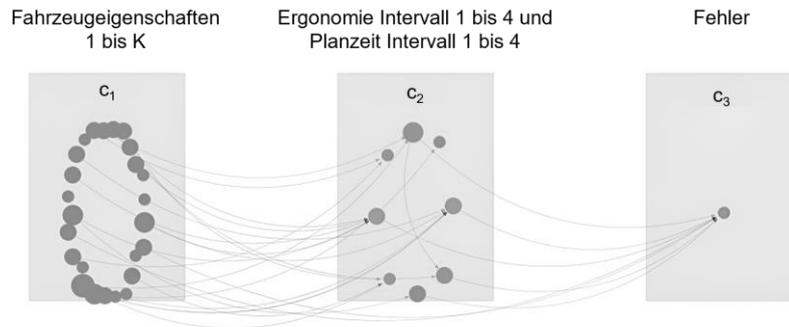

**Abb. 6.** Angenommene kausale Struktur der vorliegenden Daten.

Die Fahrzeugeigenschaften 1 bis $K$ umfassen $K = 670$ der $N = 679$ Attribute. Die Anzahl möglicher DAG wächst exponentiell mit der Anzahl der Knoten [21]. Um ihre Anzahl zu reduzieren und damit den Lösungsraum einzuschränken, wurde der Phi-Korrelationskoeffizient paarweise zwischen allen Attributen 1 bis $K$ und jeweils der Ergonomie Intervall 1 bis 4 sowie der Planzeit Intervall 1 bis 4 berechnet. Damit ergeben sich pro Attribut FaE 8 Phi-Korrelationskoeffizienten. Der Phi-Korrelationskoeffizient ist ein Maß für die Korrelation zwischen binären Attributen. Um diesen zu berechnen, betrachtet man die Kontingenztabelle der binären Attribute $X$ und $Y$. Diese ist Tabelle 4 dargestellt. Die Werte $n_{ij}$ sind die Anzahlen der jeweiligen Elemente, $n_{*j}$ ist die Spaltensumme der Spalte $j$, $n_{*0}$ ist die Zeilensumme der Zeile $i$ und $n$ ist die Gesamtanzahl aller Elemente.

**Tabelle 4.** Kontingenztabelle der binären Attribute $X$ und $Y$.

|  | $y = 0$ | $y = 1$ | Gesamt |
|---|---|---|---|
| $x = 0$ | $n_{00}$ | $n_{01}$ | $n_{0*}$ |
| $x = 1$ | $n_{10}$ | $n_{11}$ | $n_{1*}$ |
| Gesamt | $n_{*0}$ | $n_{*1}$ | $n$ |

Der Phi-Korrelationskoeffizient wird mittels der Werte der Kontingenztabelle berechnet [39]:

$$\phi = \frac{n_{11}n_{00} - n_{10}n_{01}}{\sqrt{n_{11}n_{00}n_{01}n_{10}}} \qquad (6)$$

Es werden Attribute FaE bei denen alle $|\phi| < 0{,}7$ verworfen. Damit reduziert sich die Anzahl der Fahrzeugeigenschaften von $K = 670$ auf $K = 25$ und die Anzahl der Attribute auf von $N = 679$ auf $N = 34$.

Mittels dieser Modellierung sollen die Ursache-Wirk-Zusammenhänge von durch Mitarbeiter verursachten Verschraubungsfehlern beim Verbau einer Komponente durch RCA mittels CDA untersucht und identifiziert werden.

### 3.3 Simulierte Daten

Im Folgenden wird die Simulation von Daten mit gleicher Struktur wie die vorliegenden Daten sowie bekannten kausalen Zusammenhängen beschrieben. Diese sollen für einen Benchmark und zum Vergleich der vorgestellten CDA genutzt werden.

Die in der Skriptsprache Python implementierte Bibliothek *causalAssembly* [40] ermöglich das Generieren von DAG zur Modellierung von kausalen Strukturen und die Simulation entsprechender Daten. Mit dieser werden 5 Datensätze mit $N = 34$ Attributen und $M \in \{500, 1.000, 2.000, 5.000, 10.000, 20.000, 50.000, 100.000\}$ Tupeln simuliert.

Für die Simulation von Daten mit gleicher Struktur wie die vorliegenden Daten wird zunächst eine kausale Struktur von Knoten $V$ und Kanten $E$ definiert. Diese besteht aus drei aufeinanderfolgende Subgraphen $C = (c_1, c_2, c_3)$. Die Knoten $V_{c_i} = (v_1, .. v_{O_{c_i}})$ haben jeweils vorwärts gerichteten Kanten $e = v_i \rightarrow v_j$ mit $v_i \in c_i$, $v_j \in c_j$ und $j > i$. Die Struktur wird in Tabelle 5 beschrieben.

**Tabelle 5.** Kausale Struktur für die Simulation von Daten mit gleicher Struktur wie die vorliegenden Daten.

| Subgraph $C$ | Anzahl Knoten $O_{C_i}$ | Attribut |
|---|---|---|
| $c_1$ | 25 | FaE |
| $c_2$ | 8 | Er, Pz |
| $c_3$ | 1 | Fe |

In der Simulation werden binäre Daten gesampelt, linear kombiniert und mittels einer nichtlinearen Funktion wieder auf binäre Daten abgebildet. Der DAG der simulierten Daten sowie eine Heat Map von dessen Adjazenzmatrix sind in Abb. 7 dargestellt. Eine Heat Map ist ein Raster, in dem jedem jedes Feld das Element einer Matrix repräsentiert. Der Wert des Elements ist durch die Farbe des Feldes kodiert. Die simulierten Daten werden im Folgenden als Ground Truth bezeichnet.

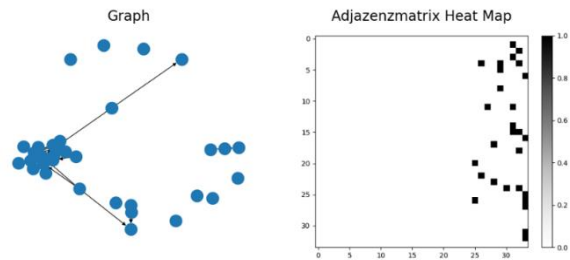

**Abb. 7.** Kausale Struktur der Ground Truth. Links: DAG, Rechts: Heat Map der Adjazenzmatrix.

### 3.4 Betrachtete Metriken

Für den Vergleich der CDA hinsichtlich der gelernten kausalen Struktur werden die gelernte und die echte Adjazenzmatrix elementweise verglichen. Sind beide Einträge 1, dann ist wird das Element als korrekt positiv (engl. True positive, Abk. TP) bezeichnet; sind beide Einträge 0, dann wird das Element als ein korrekt negativ (engl. True negative; Abk. TN) bezeichnet; ist der Eintrag der gelernten Adjazenzmatrix 1 und der echten Adjazenzmatrix 0, dann wird das Element als falsch positiv (engl. False positive,

Abk. FP) bezeichnet; ist der Eintrag der gelernten Adjazenzmatrix 0 und der echten Adjazenzmatrix 1, dann wird das Element als falsch negativ (engl. False negative, Abk. FN) bezeichnet. Die Anzahl der Einträge für TP, TN, FP, FN wird jeweils über alle Elemente der Matrix summiert. Die gelernte kausale Struktur wird anhand der Metriken Hamming-Abstand (engl. Structural Hamming Distance, Abk. SHD), Precision-Maß, Recall-Maß und F1-Maß (eng. F1-Score, Abk. F1) verglichen. Der Hamming-Abstand ergibt sich zu:

$$\text{SHD} = \text{FP} + \text{FN} \tag{7}$$

Dies entspricht der Anzahl an Kanten der gelernten Adjazenzmatrix, die nicht mit der echten Adjazenzmatrix übereinstimmen. Der ideale Wert ist 0. Das Precision-Maß ergibt sich zu:

$$\text{Precision} = \frac{\text{TP}}{\text{TP} + \text{FP}} \tag{8}$$

Es gibt an, wie viele der gelernten Kanten korrekt sind. Der ideale Wert ist 1. Das Recall-Maß ergibt sich zu:

$$\text{Recall} = \frac{\text{TP}}{\text{TP} + \text{FN}} \tag{9}$$

Es gibt an, wie viele der korrekten Kanten gelernt werden. Der ideale Wert ist 1. Precision-Maß und Recall-Maß werden auch im harmonischen Mittel als F1-Maß zusammengefasst. Dies ergibt sich zu:

$$\text{F1} = \frac{2\text{TP}}{2\text{TP} + \text{FP} + \text{FN}} \tag{10}$$

Mit diesem werden sowohl falsch positive als auch falsch negative Einträge berücksichtigt. Der ideale Wert ist 1.

## 4   Ergebnisse

Die in Kapitel 3.1 vorgestellten CDA werden für einen Benchmark auf die in Kapitel 3.3 beschriebenen simulierten Daten angewandt. Diese haben die gleiche Struktur wie die in Kapitel 3.2 beschriebenen vorliegenden Daten sowie bekannte kausale Zusammenhänge. Sie umfassen 8 Datensätze mit jeweils $N = 34$ Attributen und $M \in \{500, 1.000, 2.000, 5.000, 10.000, 20.000, 50.000, 100.000\}$ Tupeln. Die CDA werden in Kapitel 4.1 in Bezug auf die gelernte kausale Struktur anhand der in Kapitel 3.4 beschriebenen Metriken sowie in Kapitel 4.2 anhand ihrer Laufzeit verglichen.

### 4.1   Kausale Struktur

In Abb. 8 sind Heat Maps der Adjazenzmatrizen der mittels der vorgestellten CDA gelernten kausalen Strukturen exemplarisch für einen verwendeten Datensatz mit $M = 100.000$ Tupeln dargestellt.

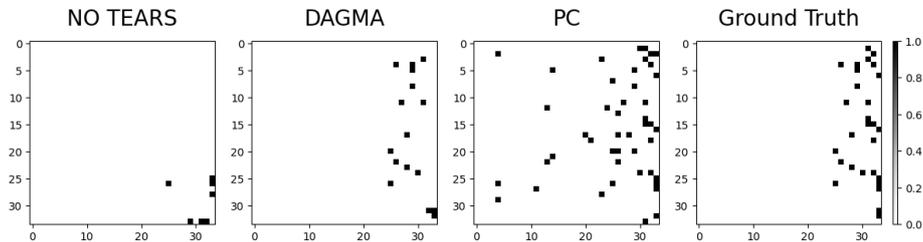

**Abb. 8.** Heat Maps der Adjazenzmatrizen der mittels der vorgestellten CDA gelernten kausalen Strukturen mit $M = 100.000$. Links: NO TEARS-Algorithmus. Mitte links: DAGMA-Algorithmus. Mitte rechts: PC-Algorithmus. Rechts: Ground Truth.

Man sieht, dass der NO TEARS-Algorithmus zu wenig Kanten in Bezug auf die Ground Truth und der PC-Algorithmus zu viele Kanten in Bezug auf die Ground Truth lernt.

In Abb. 9 bis Abb. 12 sind die beschriebenen Metriken der mittels der vorgestellten CDA gelernten kausalen Strukturen in Abhängigkeit der Anzahl $M = \{500, 1.000, 2.000, 5.000, 10.000, 20.000, 50.0000, 100.000\}$ der Tupel der verwendeten Datensätze dargestellt. Die Metriken wurden jeweils anhand von 25 simulierten Datensätzen berechnet. In den Plots ist jeweils der arithmetische Mittelwert der Metrik und in den Tabellen der arithmetische Mittelwert der Metrik sowie ihre Standardabweichung dargestellt.

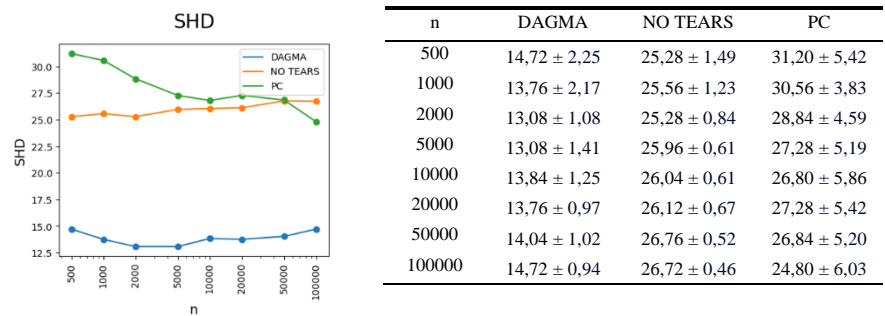

| n | DAGMA | NO TEARS | PC |
| --- | --- | --- | --- |
| 500 | 14,72 ± 2,25 | 25,28 ± 1,49 | 31,20 ± 5,42 |
| 1000 | 13,76 ± 2,17 | 25,56 ± 1,23 | 30,56 ± 3,83 |
| 2000 | 13,08 ± 1,08 | 25,28 ± 0,84 | 28,84 ± 4,59 |
| 5000 | 13,08 ± 1,41 | 25,96 ± 0,61 | 27,28 ± 5,19 |
| 10000 | 13,84 ± 1,25 | 26,04 ± 0,61 | 26,80 ± 5,86 |
| 20000 | 13,76 ± 0,97 | 26,12 ± 0,67 | 27,28 ± 5,42 |
| 50000 | 14,04 ± 1,02 | 26,76 ± 0,52 | 26,84 ± 5,20 |
| 100000 | 14,72 ± 0,94 | 26,72 ± 0,46 | 24,80 ± 6,03 |

**Abb. 9.** Hamming-Distanz der gelernten kausalen Strukturen in Abhängigkeit von der Anzahl der Tupel der verwendeten Datensätze. Links: Plot des arithmetischen Mittelwertes. Rechts: Arithmetischer Mittelwert und Standardabweichung.

Der DAGMA-Algorithmus weist die geringste Hamming-Distanz zur Ground Truth auf. Diese ist nicht von der Anzahl der Tupel der verwendeten Datensätze abhängig. Für den NO TEARS-Algorithmus und den PC-Algorithmus ergibt sich eine vergleichbare Hamming-Distanz zur Ground Truth. Für den PC-Algorithmus verringert sich diese für eine wachsende Anzahl der Tupel der verwendeten Datensätze. Für den NO TEARS-Algorithmus ist diese nicht von Anzahl der Tupel der verwendeten Datensätze abhängig.

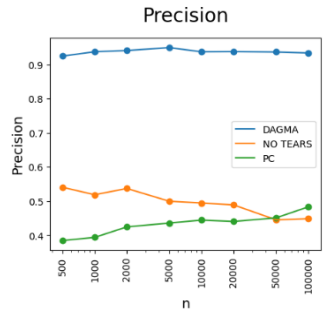

| n | DAGMA | NO TEARS | PC |
|---|---|---|---|
| 500 | 0,93 ± 0,05 | 0,54 ± 0,089 | 0,38 ± 0,08 |
| 1000 | 0,94 ± 0,04 | 0,51 ± 0,07 | 0,39 ± 0,06 |
| 2000 | 0,94 ± 0,02 | 0,54 ± 0,05 | 0,42 ± 0,06 |
| 5000 | 0,95 ± 0,03 | 0,50 ± 0,04 | 0,43 ± 0,06 |
| 10000 | 0,94 ± 0,05 | 0,49 ± 0,04 | 0,44 ± 0,06 |
| 20000 | 0,94 ± 0,04 | 0,49 ± 0,04 | 0,44 ± 0,07 |
| 50000 | 0,94 ± 0,01 | 0,45 ± 0,04 | 0,45 ± 0,07 |
| 100000 | 0,93 ± 0,01 | 0,45 ± 0,03 | 0,48 ± 0,08 |

**Abb. 10.** Precision-Maß der gelernten kausalen Strukturen in Abhängigkeit von der Anzahl der Tupel der verwendeten Datensätze. Links: Plot des arithmetischen Mittelwertes. Rechts: Arithmetischer Mittelwert und Standardabweichung.

Der DAGMA-Algorithmus weist das höchste Precision-Maß auf. Dieses ist nicht von der Anzahl der Tupel der verwendeten Datensätze abhängig. Für den NO TEARS-Algorithmus und den PC-Algorithmus ergeben sich hinsichtlich des Precision-Maßes vergleichbare Werte. Für den PC-Algorithmus erhöht sich das Precision-Maß für eine wachsende Anzahl der Tupel der verwendeten Datensätze. Für den NO TEARS-Algorithmus verringert sich das Precision-Maß für eine wachsende Anzahl der Tupel der verwendeten Datensätze. Das Precision-Maß gibt an, wie viele der gelernten Kanten korrekt sind. Da die mittels des PC-Algorithmus gelernte kausale Struktur mehr Kanten besitzt, weist diese trotz vergleichbaren Precision-Maßes mehr korrekte Kanten auf.

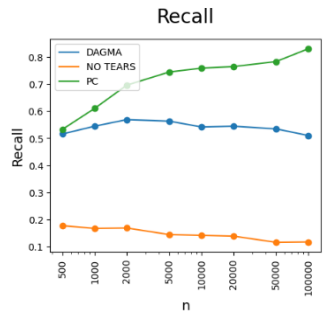

| n | DAGMA | NO TEARS | PC |
|---|---|---|---|
| 500 | 0,52 ± 0,07 | 0,18 ± 0,03 | 0,53 ± 0,09 |
| 1000 | 0,54 ± 0,06 | 0,017 ± 0,03 | 0,71 ± 0,07 |
| 2000 | 0,57 ± 0,04 | 0,17 ± 0,03 | 0,70 ± 0,08 |
| 5000 | 0,56 ± 0,05 | 0,14 ± 0,02 | 0,74 ± 0,08 |
| 10000 | 0,54 ± 0,04 | 0,14 ± 0,02 | 0,76 ± 0,08 |
| 20000 | 0,54 ± 0,03 | 0,14 ± 0,02 | 0,76 ± 0,08 |
| 50000 | 0,53 ± 0,04 | 0,12 ± 0,02 | 0,78 ± 0,07 |
| 100000 | 0,51 ± 0,03 | 0,12 ± 0,02 | 0,83 ± 0,08 |

**Abb. 11.** Recall-Maß der gelernten kausalen Strukturen in Abhängigkeit von der Anzahl der Tupel der verwendeten Datensätze. Links: Plot des arithmetischen Mittelwertes. Rechts: Arithmetischer Mittelwert und Standardabweichung.

Der PC-Algorithmus weist das höchste Recall-Maß auf. Dieses erhöht sich für eine wachsende Anzahl der Tupel der verwendeten Datensätze. Für den NO TEARS-Algorithmus ergibt sich hinsichtlich des Recall-Maßes ein geringerer Werte und der Wert für den NO TEARS-Algorithmus am geringsten. Das Recall-Maß für den DAGMA-Algorithmus und den NO TEARS-Algorithmus ist nicht von der Anzahl der Tupel der verwendeten Datensätze abhängig. Das Recall-Maß gibt an, wie viele der korrekten Kanten gelernt werden. Da die mittels des PC-Algorithmus gelernte kausale Struktur die meisten Kanten besitzt, weist diese die meisten korrekten Kanten auf.

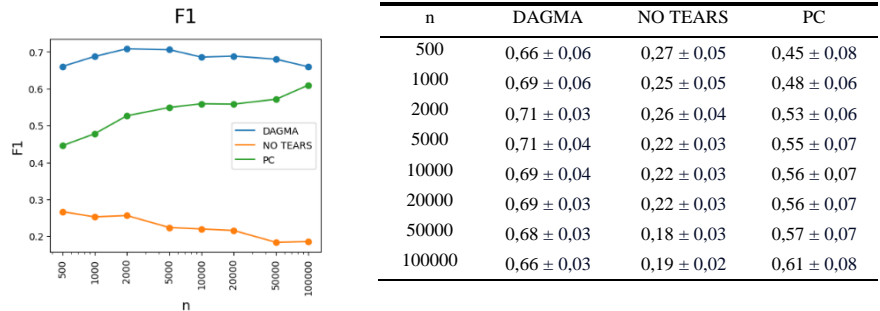

| n | DAGMA | NO TEARS | PC |
|---|---|---|---|
| 500 | 0,66 ± 0,06 | 0,27 ± 0,05 | 0,45 ± 0,08 |
| 1000 | 0,69 ± 0,06 | 0,25 ± 0,05 | 0,48 ± 0,06 |
| 2000 | 0,71 ± 0,03 | 0,26 ± 0,04 | 0,53 ± 0,06 |
| 5000 | 0,71 ± 0,04 | 0,22 ± 0,03 | 0,55 ± 0,07 |
| 10000 | 0,69 ± 0,04 | 0,22 ± 0,03 | 0,56 ± 0,07 |
| 20000 | 0,69 ± 0,03 | 0,22 ± 0,03 | 0,56 ± 0,07 |
| 50000 | 0,68 ± 0,03 | 0,18 ± 0,03 | 0,57 ± 0,07 |
| 100000 | 0,66 ± 0,03 | 0,19 ± 0,02 | 0,61 ± 0,08 |

**Abb. 12.** F1-Maß der gelernten kausalen Strukturen in Abhängigkeit von der Anzahl der Tupel der verwendeten Datensätze. Links: Plot des arithmetischen Mittelwertes. Rechts: Arithmetischer Mittelwert und Standardabweichung.

Das F1-Maß ist das harmonische Mittel zwischen dem Precision-Maß und dem Recall-Maß. Der DAGMA-Algorithmus weist das höchste Precision-Maß auf. Für den PC-Algorithmus ergibt sich das höchste Recall-Maß. Damit weisen diese beiden Algorithmen ein höheres F1-Maß als der NO TEARS-Algorithmus auf. Das F1-Maß für den DAGMA-Algorithmus ist dabei am höchsten. Dieses ist nicht von der Anzahl der Tupel der verwendeten Datensätze abhängig. Das F1-Maß für den PC-Algorithmus erhöht sich für eine wachsende Anzahl der Tupel der verwendeten Datensätze. Das F1-Maß für den NO TEARS-Algorithmus ist nicht von der Anzahl der Tupel der verwendeten Datensätze abhängig.

### 4.2 Laufzeit

In Abb. 13 ist die Laufzeit der vorgestellten CDA in Abhängigkeit der Anzahl $M = \{500, 1.000, 2.000, 5.000, 10.000, 20.000, 50.000, 100.000\}$ der Tupel der verwendeten Datensätze dargestellt. Die Laufzeiten des PC-Algorithmus und des NO TEARS-Algorithmus werden mittels ihrer Implementierungen in Bibliothek *gCastle* [41] ermittelt. Die Laufzeiten des DAGMA-Algorithmus wird mittels ihrer Implementierung in der Bibliothek *dagma* [37] ermittelt. Die Bibliotheken *gCastle* und *dagma* sind in der Skriptsprache Python implementiert.

Die Laufzeit der CDA wurde auf einer AWS EC2 g3.4xlarge Instanz jeweils 25 Mal gemessen. In dem Plot ist der arithmetische Mittelwert der Laufzeiten und in den Tabellen der arithmetische Mittelwert der Laufzeit sowie ihre Standardabweichung dargestellt.

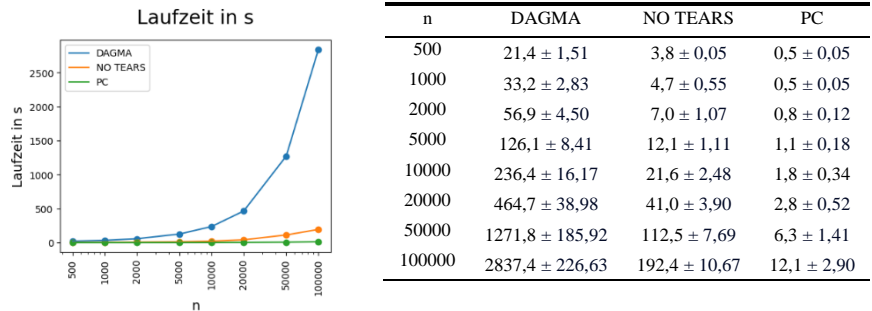

| n | DAGMA | NO TEARS | PC |
| --- | --- | --- | --- |
| 500 | 21,4 ± 1,51 | 3,8 ± 0,05 | 0,5 ± 0,05 |
| 1000 | 33,2 ± 2,83 | 4,7 ± 0,55 | 0,5 ± 0,05 |
| 2000 | 56,9 ± 4,50 | 7,0 ± 1,07 | 0,8 ± 0,12 |
| 5000 | 126,1 ± 8,41 | 12,1 ± 1,11 | 1,1 ± 0,18 |
| 10000 | 236,4 ± 16,17 | 21,6 ± 2,48 | 1,8 ± 0,34 |
| 20000 | 464,7 ± 38,98 | 41,0 ± 3,90 | 2,8 ± 0,52 |
| 50000 | 1271,8 ± 185,92 | 112,5 ± 7,69 | 6,3 ± 1,41 |
| 100000 | 2837,4 ± 226,63 | 192,4 ± 10,67 | 12,1 ± 2,90 |

**Abb. 13.** Laufzeit der vorgestellten CDA in Abhängigkeit der Anzahl $M = \{500, 1.000, 2.000, 5.000, 10.000, 20.000, 50.000, 100.000\}$ der Tupel der verwendeten Datensätze. Links: Plot des arithmetischen Mittelwertes. Rechts: Arithmetischer Mittelwert und Standardabweichung.

Man sieht, dass die Laufzeit aller betrachteten CDA mit steigender Anzahl der Tupel der verwendeten Daten annähernd exponentiell wächst. Der DAGMA-Algorithmus hat dabei eine bis über zwölffach längere Laufzeit im Vergleich zum NO TEARS-Algorithmus und zum PC-Algorithmus.

## 5 Diskussion

In dieser Veröffentlichung soll untersucht werden, wie RCA in der Montage von Fahrzeugen mittels CDA durchführt werden kann. Dazu werden CDA vorgestellt und die Verknüpfung und Vorverarbeitung vorliegender Daten beschrieben. Anschließend werden die vorgestellten CDA auf simulierte Daten mit gleicher Struktur wie die vorliegenden Daten sowie bekannten kausalen Zusammenhängen angewandt. Danach werden diese anhand der gelernten kausalen Struktur mittels verschiedener Metriken sowie ihrer Laufzeit miteinander verglichen.

Der DAGMA-Algorithmus liefert in Bezug auf die Metriken Hamming-Distanz, Precision- sowie F1-Maß die besten Ergebnisse. Die Metriken hängen nicht von der Anzahl der Tupel der verwendeten Daten ab. Gleichzeitig ist die Laufzeit von diesem Algorithmus am höchsten. Daraus ergibt sich, dass der DAGMA-Algorithmus zur RCA verwendet werden sollte, wenn die Laufzeit zu vernachlässigen ist.

Der PC-Algorithmus liefert in Bezug auf die Metrik Recall-Maß das beste Ergebnis. Gleichzeitig ist die Laufzeit von diesem Algorithmus am niedrigsten. Die Metriken verbessern sich mit wachsender Anzahl der Tupel der verwendeten Daten. Außerdem lernt der PC-Algorithmus die meisten korrekten Kanten. Die nicht korrekt gelernte Kanten könnten durch das Einbeziehen von domänenspezifischem Fachwissen identifiziert werden. Daraus ergibt sich, dass der PC-Algorithmus bei großen Mengen an vorliegenden Daten und domänenspezifischem Fachwissen sowie bei Einschränkungen hinsichtlich der Laufzeit zur RCA verwendet werden sollte.

Der NO TEARS-Algorithmus lieferte in Bezug auf Hamming-Distanz und Precision ähnliche Werte wie der PC-Algorithmus. In Bezug auf Recall-Maß und F1-Maß liefert dieser die schlechtesten Ergebnisse. Die Metriken hängen nicht von der Anzahl der Tupel der verwendeten Daten ab oder verschlechtern sich wachsender Anzahl der Tupel der verwendeten Daten. Daraus ergibt sich, dass von der Verwendung des NO TEARS-Algorithmus abgesehen werden sollte.